\def\year{2022}\relax
\documentclass[letterpaper]{article} 
\usepackage{aaai22}  
\usepackage{times}  
\usepackage{helvet}  
\usepackage{courier}  
\usepackage[hyphens]{url}  
\usepackage{graphicx} 
\usepackage{natbib}  
\usepackage{caption} 
\DeclareCaptionStyle{ruled}{labelfont=normalfont,labelsep=colon,strut=off} 
\usepackage{algorithm}

\usepackage{amsmath}
\usepackage{bm}
\usepackage{amssymb}
\usepackage{booktabs}
\usepackage{multirow}

\usepackage{algorithmicx}
\usepackage{algpseudocode}
\usepackage{xcolor}

\usepackage{newfloat}
\usepackage{listings}
\floatstyle{ruled}
\newfloat{listing}{tb}{lst}{}
\floatname{listing}{Listing}
    \title{Channelized Axial Attention -- \\ Considering Channel Relation within Spatial Attention for Semantic Segmentation}
\author {
    Ye Huang\textsuperscript{\rm 1}\quad
    Di Kang \textsuperscript{\rm 2}\quad
    Wenjing Jia\textsuperscript{\rm 1}\quad
    Xiangjian He\textsuperscript{\rm 1}\quad
    Liu Liu\textsuperscript{\rm 3}\\
}
\affiliations {
    \textsuperscript{\rm 1} University of Technology Sydney\quad
    \textsuperscript{\rm 2} Tencent AI LAB\quad
    \textsuperscript{\rm 3} University of Sydney \\
    edward.ye.huang@qq.com
}

\usepackage{bibentry}

\begin{document}
\copyrighttext{Under Review}
\maketitle
\begin{abstract}
Spatial and channel attentions, modelling the semantic interdependencies in spatial and channel dimensions respectively, have recently been widely used for semantic segmentation.
However, computing spatial and channel attentions separately sometimes causes errors, especially for those difficult cases.
In this paper, we propose Channelized Axial Attention (CAA) to seamlessly \textit{integrate} channel attention and spatial attention into a single operation with negligible computation overhead. 
Specifically, we break down the dot-product operation of the spatial attention into two parts and insert channel relation in between, allowing for independently optimized channel attention on each spatial location.
We further develop grouped vectorization, which allows our model to run with very little memory consumption without slowing down the running speed.
Comparative experiments conducted on multiple benchmark datasets, including Cityscapes, PASCAL Context, and COCO-Stuff, demonstrate that our CAA  outperforms many state-of-the-art  segmentation  models (including dual attention) on all tested datasets.
\end{abstract}

\begin{figure}[!ht]
	\centering
	\includegraphics[keepaspectratio=true, width=0.98\linewidth]{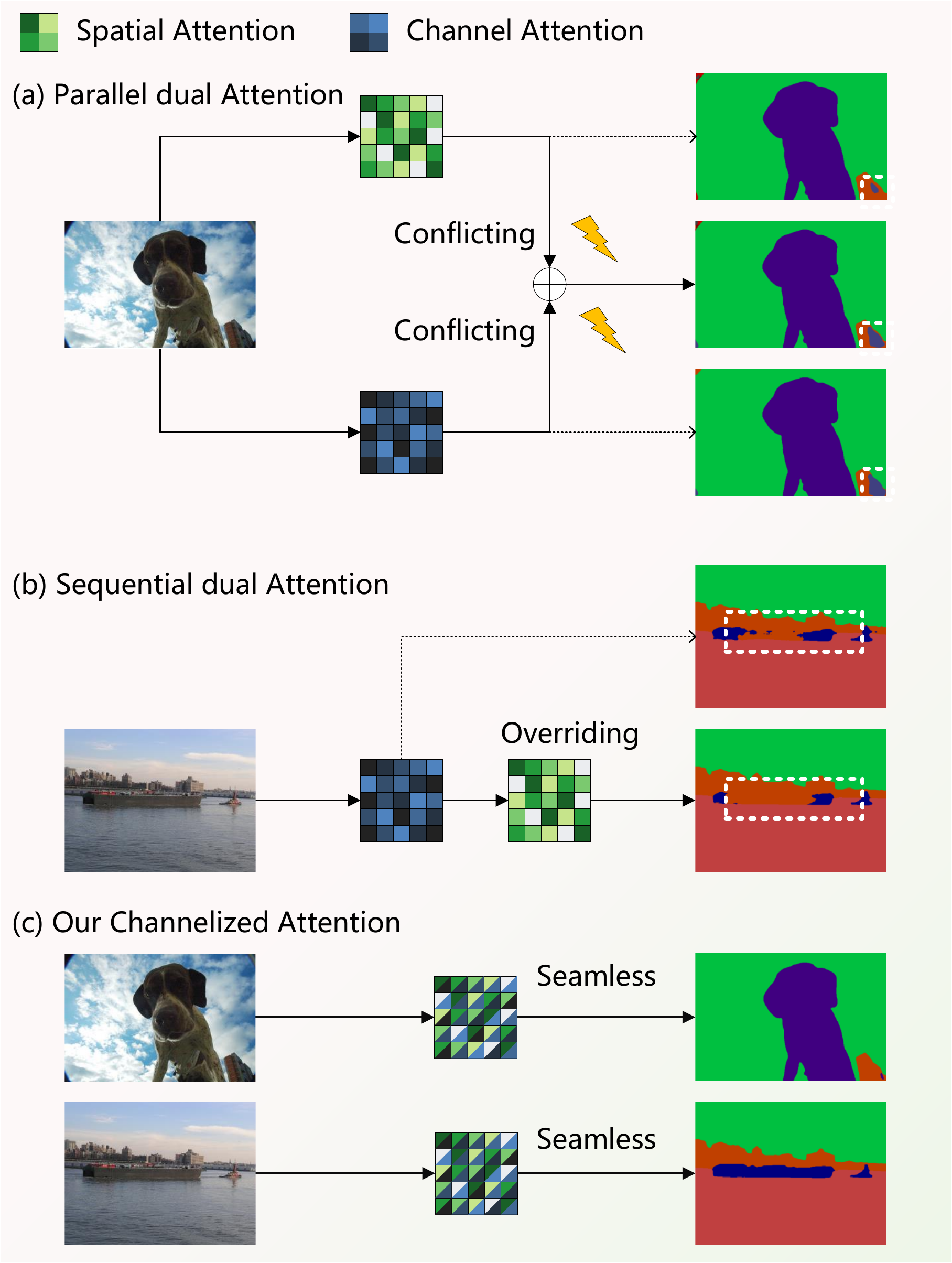}
	\caption{Different dual attention designs: (a) \textbf{Parallel dual attention} sums the results from spatial and channel attentions directly, which may cause conflicts because spatial and channel attentions are focusing on different aspects. (b) \textbf{Sequential dual attention} performs spatial attention after channel attention, where the spatial attention may override correct features extracted by the channel attention. (c) \textbf{Our channelized attention} seamlessly merges the spatial and channel attentions into a single operation (see Sect.~\ref{sChannelizedAttention}), removing the potential conflicting issue caused by \textbf{a} or \textbf{b}.}
	\label{fig0}
\end{figure}

\section{Introduction}
\label {sIntroduction}

Semantic segmentation is a fundamental task in many computer vision applications, which assigns a class label to each pixel in the image. 
Most of the existing approaches~\cite{cOCR,cDenseASPP,cDualAttention,cEMANet} have adopted a pipeline similar to the one that is defined by Fully Convolutional Networks (FCNs)~\cite{cFCN} using fully convolutional layers to output the pixel-level segmentation results of input images. These approaches have achieved state-of-the-art performance. 
After the FCNs,
there have been many approaches dedicated to extracting enhanced pixel representations from the backbone. 
Earlier approaches, including PSPNet~\cite{cPSPNet} and DeepLab~\cite{cDeepLabV3Plus}, used a Pyramid Pooling Module or an Atrous Spatial Pyramid Pooling module to expand the receptive field to enhance the representation capabilities. 
Recently, many works focus on using the attention mechanisms to enhance pixel representations. 
The first attempts in this direction included Squeeze and Excitation Networks (SENets)~\cite{cSENet} that introduced a simple yet effective channel attention module to explicitly model the interdependencies between channels. 
Meanwhile, spatial attention relied on self-attention proposed in~\cite{cNonLocal, cAttentionIsAllYourNeed} to model long-range dependencies in spatial domain, so as to produce more correct pixel representations.
For each pixel in the feature maps, spatial attention ``corrects'' its representation with the representations of other pixels depending on their similarity.
In contrast, channel attention identifies important channels based on all spatial locations and reweights the extracted features.

\textit{Parallel dual attention} (\textit{e.g.},~\cite{cDualAttention}) was proposed to gain the advantages of both spatial attention and channel attention. This approach directly fused their results with an element-wise addition (see Fig.~\ref{fig0}(a)). 
Although they have achieved improved performance, the relationship between the contributions of spatial and channel attentions to the final results is unclear.
Moreover, calculating the two attentions separately not only increases the computational complexity, but may also result in conflicting importance of feature representations. 
For example, some channels may appear to be important in spatial attention for a pixel that belongs to a partial region in the feature maps. However, channel attention may have its own perspective, which is calculated by summing up the similarities over the entire feature maps, and weakens the impact of spatial attention. 

\textit{Sequential dual attention}, which combines channel attention and spatial attention in a sequential manner (Fig.~\ref{fig0}(a)) has similar issues. 
For example, channel attention can \textit{ignore} the partial region representation obtained from the overall perspective. However, this partial region representation may be required by spatial attention. 
Thus, directly fusing the spatial and channel attention results may yield incorrect importance weights for pixel representations. 
In Sect.~\ref{sExperements}, we develop an approach to visualize the impact of the conflicting feature representation on the final segmentation results. 

In order to overcome the aforementioned issues, we propose Channelized Axial Attention (CAA), which breaks down the axial attention into more basic parts and inserts channel attention into them, combining spatial attention and channel attention together seamlessly and efficiently.
Specifically, when applying the axial attention maps to the input signal~\cite{cNonLocal}, we capture the intermediate results of the dot product before they are summed up along the corresponding axes. 
Capturing these intermediate results allows channel attention to be integrated for each column and each row, instead of computing on the mean or sum of the features in the entire feature maps. 
We also develop a novel grouped vectorization approach to maximize the computation speed in limited GPU memory.

In summary, our contributions in this paper include:
\begin{itemize}
	%
    \item We are the first to explicitly \textit{identify} the potential conflicts between spatial and channel attention in existing dual attention designs by \textit{visualizing} the effects of each attention on the final result.

	\item We propose a novel Channelized Axial Attention, which breaks down the axial attention into more basic parts and inserts channel attention in between, integrating spatial attention and channel attention together seamlessly and efficiently, with only a minor computation overhead compared to the original axial attention. 
	\item To balance the computation speed and GPU memory usage, a grouped vectorization approach for computing the channelized attentions is proposed. This is particularly advantageous when processing large images.
	
	\item Experiments on three challenging benchmark datasets, including PASCAL Context~\cite{cPascalVOC}, COCO-Stuff~\cite{cCocoStuff} and Cityscapes~\cite{cCityScapes}, demonstrate the superiority of our approach over the state-of-the-art approaches.
\end{itemize}


\section{Related Work}
\label{sRelatedWorks}

\noindent\textbf{Spatial attention. }
Non-local networks~\cite{cNonLocal} and Transformer~\cite{cAttentionIsAllYourNeed} introduced the self-attention mechanism to examine the pixel relationship in the spatial domain. 
It usually calculates dot-product similarity or cosine similarity to obtain the similarity measurement between every two pixels in feature maps, and recalculates the feature representation of each pixel according to its similarity with others. 
Self-attention has successfully addressed the feature map coverage issue of multiple fixed-range approaches~\cite{cDeepLab, cPSPNet}, but it has also introduced huge computation costs for computing the complete feature map. 
This means that, for each pixel in the feature maps, its attention similarity affects all other pixels. 
Recently, many approaches~\cite{cCCNet,cANNN} have been developed to optimize the GPU memory costs of spatial self-attention.\\

\noindent\textbf{Channel Attention. }
Channel attention~\cite{cSENet} examined the relationships between channels, and enhanced the important channels so as to improve performance. 
SENets~\cite{cSENet} conducted a global average pooling to get mean feature representations, and then went through two fully connected layers, where the first one reduced channels and the second one recovered the original channels, resulting in channel-wise weights according to the importance of channels. 
In DANet~\cite{cDualAttention}, channel-wise relationships were modelled by a 2D attention matrix, similar to the self-attention used in the spatial domain, except that it computed the attention with a dimension of $C \times C$ rather than $(H \times W) \times (H \times W)$ (here, $C$ represents the number of channels, and $H$ and $W$ represent the height and width of the feature maps, respectively).\\

\noindent\textbf{Spatial Attention + Channel Attention. }
\label{secDualAttentionDesign}
Combining spatial attention and channel attention can provide fully optimized pixel representations in a feature map. 
However, it is not easy to enjoy both advantages seamlessly. 
In DANet~\cite{cDualAttention}, the results of the channel attention and spatial attention are directly added together.
Supposing that there is a pixel belonging to a semantic class that has a tiny region in the feature maps, spatial-attention can find its similar pixels. 
However, channel representation of the semantic class with a partial region of the feature maps may not be important in the perspective of entire feature maps, so it may be ignored when conducting channel attention computations. 
Computing self-attention and channel attention separately (as illustrated in Fig.~\ref{fig0}(a)) can cause conflicting results, and thus weaken their performance when both results are summarized together.  
Similarly, in the cascaded model (see Fig.~\ref{fig0}(b)), the spatial attention module after the channel attention module may pick up an incorrect pixel representation enhanced by channel attention, because channel attention computes channel importance according to the entire feature maps. 

\begin{figure}[t]
	\centering
	\includegraphics[width=0.95\linewidth]{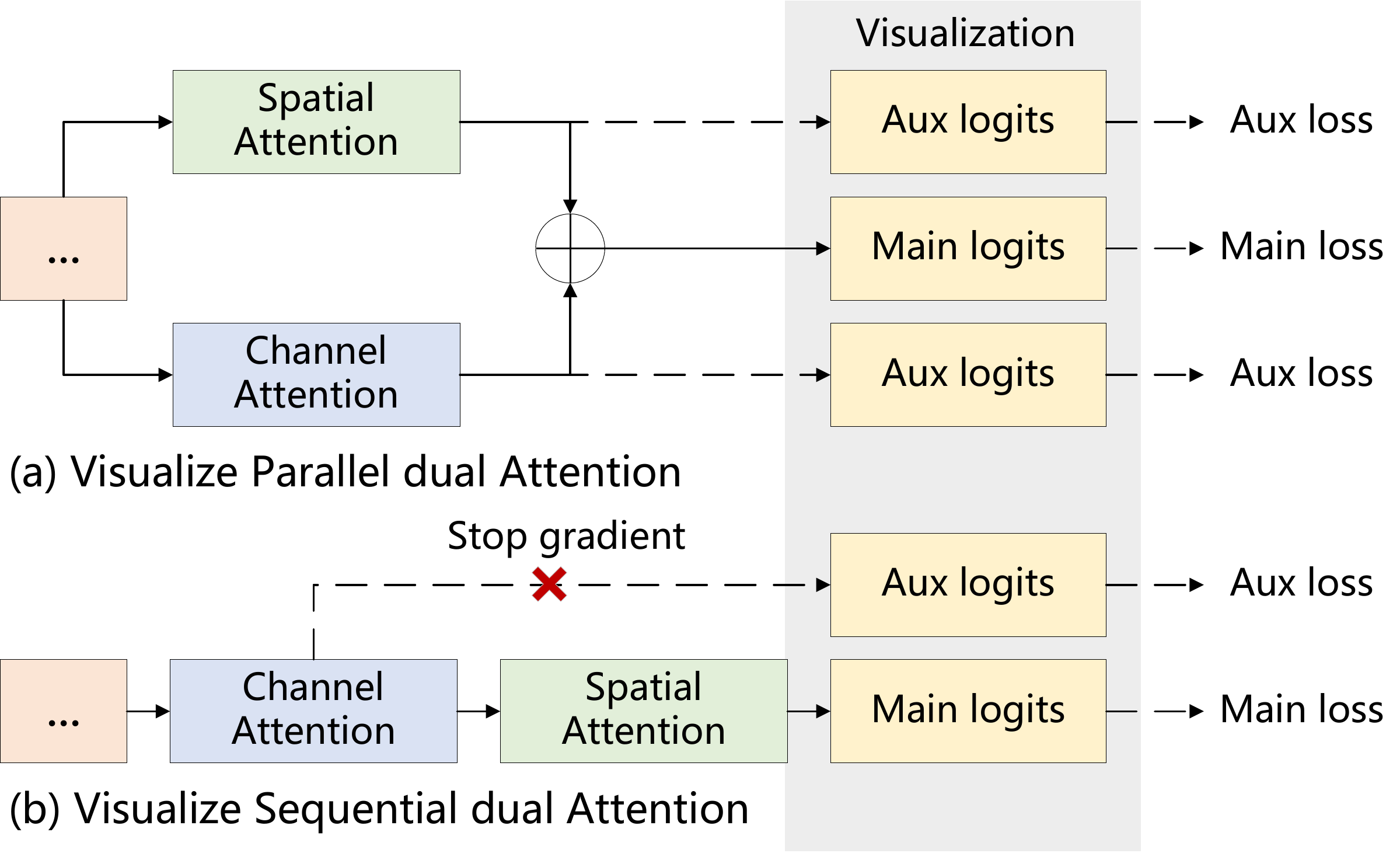}
	\caption{Our designs for visualizing the effects of dual attentions in parallel and sequential.}
	\label{fvis_cf_method}
\end{figure}

 \begin{figure}[t]
     \centering
     \includegraphics[width=1\linewidth]{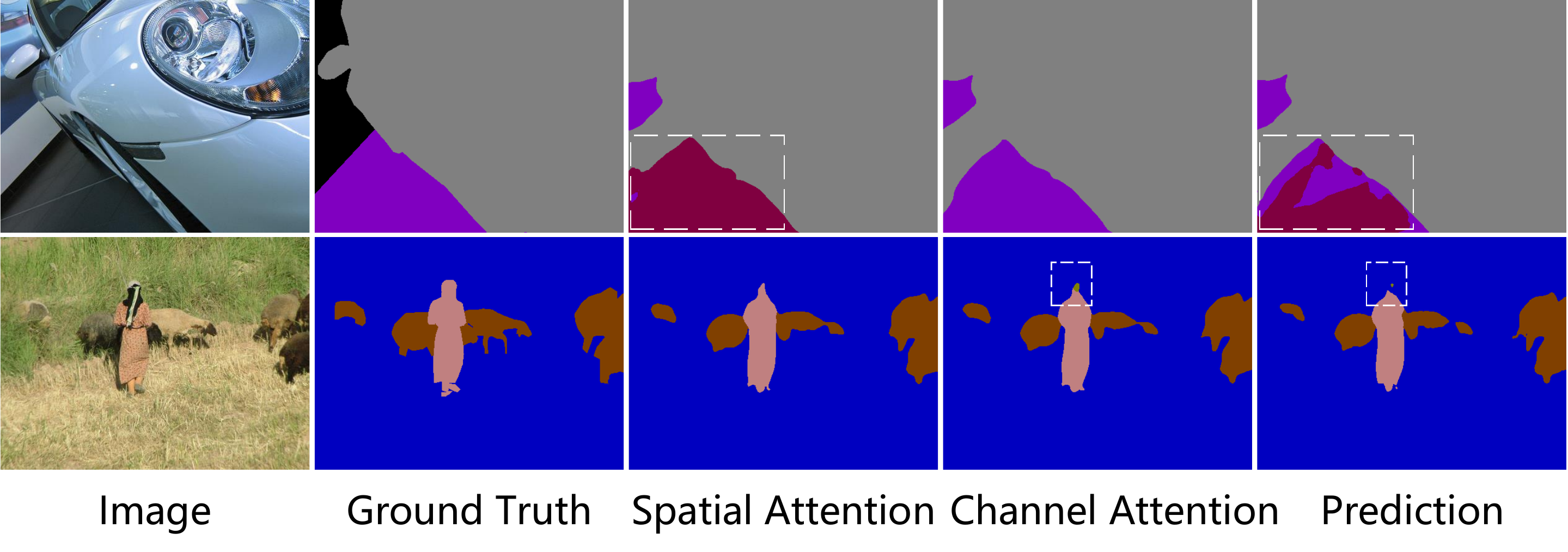}
     \caption{Conflicting features in parallel dual attention designs. \textbf{Top:} The bad spatial attention representation negatively influences the good channel attention representation. \textbf{Bottom:} The bad channel attention representation negatively influences the good spatial attention representation. See the boxed areas.}
     \label{fdanet_pam_cam}
 \end{figure}
 \hfill
 \begin{figure}[t]
     \centering
     \includegraphics[width=1\linewidth]{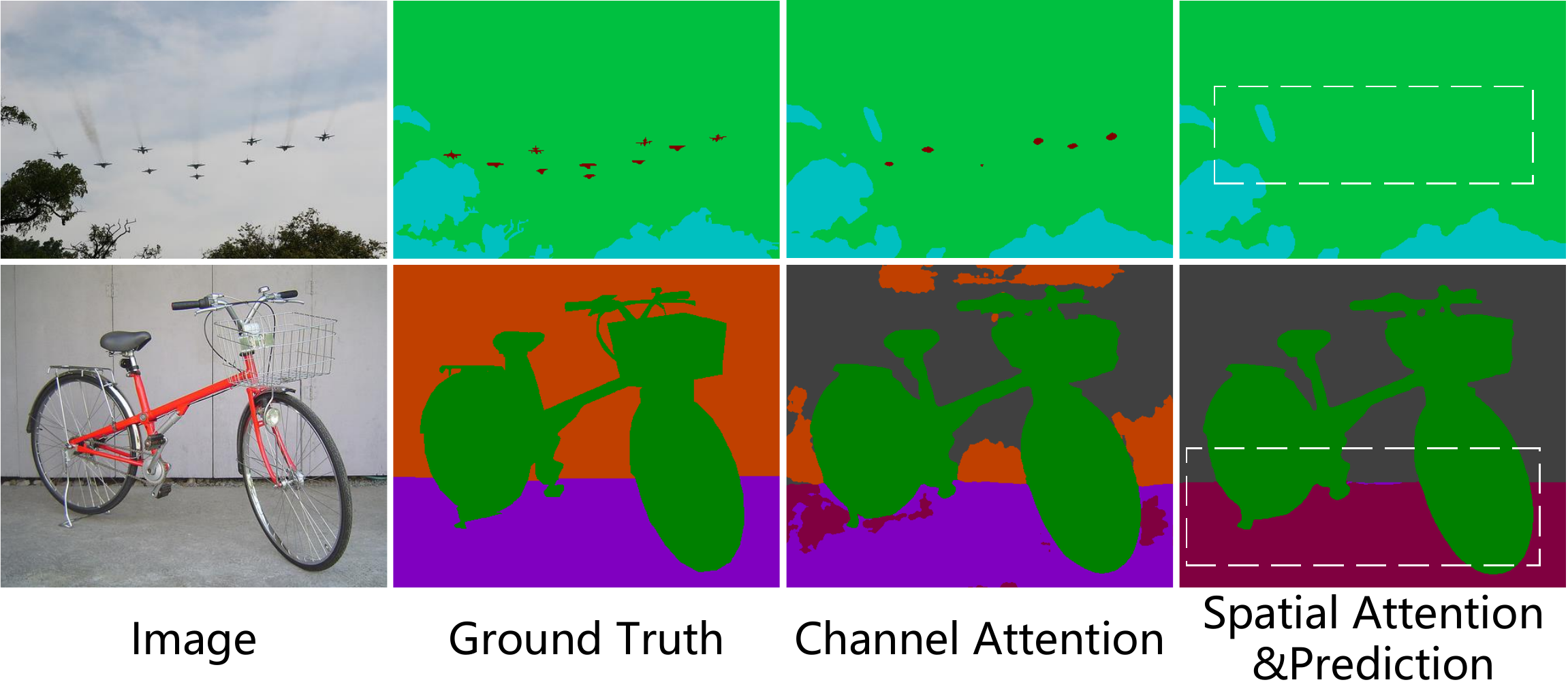}
     \caption{In sequential dual attention designs, the spatial attention representation (the 4th column) ignores the correct channel  attention representation (the 3rd column).}
     \label{fig:axialse_vs_sa}
      \label{fig:conflict}
 \end{figure}

\section{Exploring Conflicting Features}

As we have analyzed earlier in Sect.~\ref{secDualAttentionDesign}, computing spatial and channel attentions separately can cause conflicting features. 
In our experiments, to illustrate this feature conflicting issue faced by existing dual attention approaches, we designed a simple way to visualize the effects of spatial attention and channel attentions on pixel representation.

\subsection{Visualizing Conflicts}
For a parallel dual attention design such as DANet~\cite{cDualAttention}, since it has two auxiliary losses for each of spatial attention and channel attention, we directly use their logits during inference to generate corresponding segmentation results and compare them with the result generated by the main logits.
For a sequential dual attention design, we add an extra branch that directly uses the pixel representation obtained from channel attention to perform the segmentation logits. 
Note that, since the original sequential design does not have independent logits after the channel attention module, we stop the gradient from back-propagating to the main branch, to ensure that our newly added branch has no effect on the main branch.

\subsection{Examples of Conflicting Features}

To visualize the impact of the feature conflicting issue in the existing dual attention designs (see Sect.~\ref{secDualAttentionDesign}), we present examples of the segmentation results obtained with the conflicting features in the parallel dual attention design (see Fig.~\ref{fdanet_pam_cam}) and the sequential dual attention design (see Fig.~\ref{fig:axialse_vs_sa}). 
As observed from Fig.~\ref{fdanet_pam_cam}, the parallel design of dual attention directly sums up the pixel representations obtained from spatial attention and channel attention. 
With this approach, the advantages of the pixel representations obtained from one can be weakened by the other. 

The sequential way of combining the dual attentions avoids taking their average but still has its own problem. 
As shown in Fig.~\ref{fig:axialse_vs_sa}, the pixel representation obtained from the spatial attention ignores the correct representation obtained from the channel attention, and worsens the prediction.


\section{Methods} 
\label {sectMethods}

\begin{figure*}[t]
	\centering
	\includegraphics[width=1.0\linewidth]{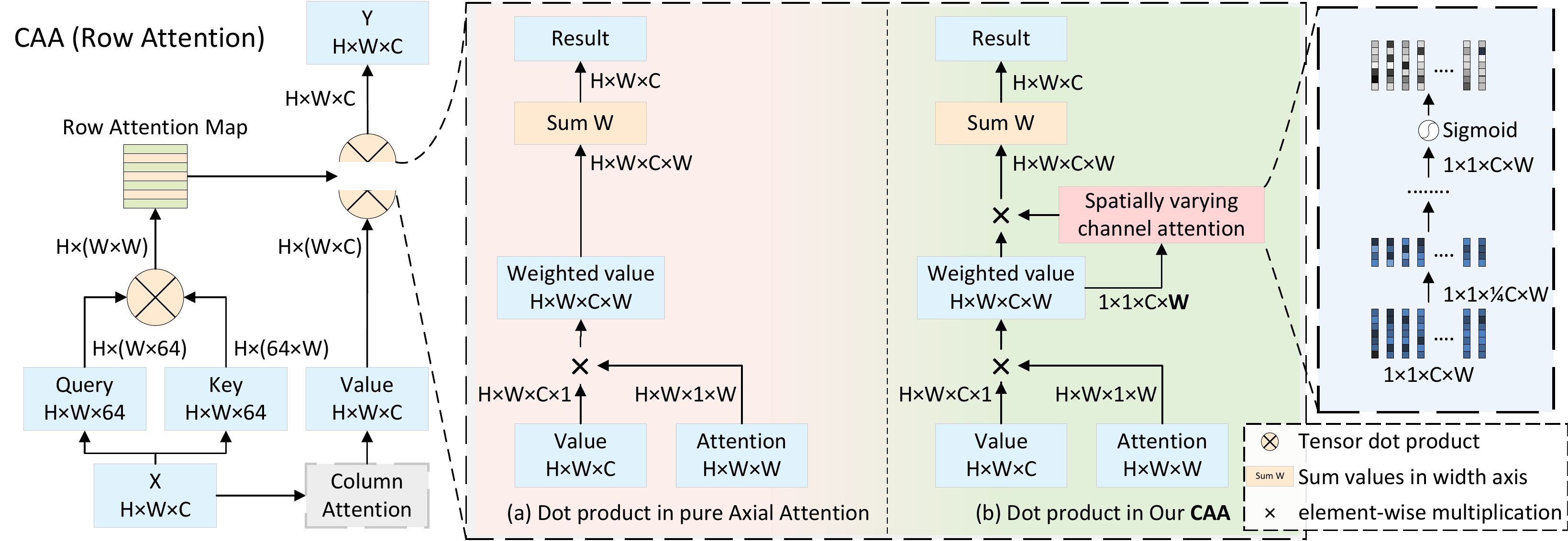}
	\caption{The detailed architecture of the proposed CAA (Row Attention). We present the way to apply channel attention seamlessly in (\textbf{b}). We mark the independent spatial dimension in \textbf{bold} style. This allows channel attention to also consider spatial unique information. \textit{Note that}, in our design, the ``\textit{value}" for row attention is obtained from the result of column attention. See Eq.~\ref{eq:CAAFinal} for details, and the \textbf{Appendix} for the full architecture.
	}
	\label{figOverall}
\end{figure*}

\subsection {Preliminaries}
\subsubsection{Formulation of the Spatial Self-attention}
Following Non Local~\cite{cNonLocal} and Stand Alone Self Attention ~\cite{cStandAloneSelfAttention}, a 2D self-attention operation in spatial domain can be defined by:
\begin{equation}
\scriptsize
\mathbf y_{i,j} = \sum_{\forall m,n} f( \mathbf{x}_{i,j},  \mathbf{x}_{m,n}) g( \mathbf{x}_{m,n}). 
\label{eq:fullattentionapply}
\end{equation}
Here, a pairwise function $f$ computes the similarity between the pixel representation $\mathbf{x}_{i,j}$ (\textit{query}) at position $(i,j)$ and the pixel representation $\mathbf{x}_{m,n}$ (\textit{key}) at all other possible positions $(m,n)$.
The unary function $g$ 
maps the original representation at position $(m,n)$ to a new domain (\textit{value}). 
In our work, we use the similarity function~\cite{cNonLocal} as $f$, \textit{i.e.},
\begin {equation}
\scriptsize
f( \mathbf{x}_{i,j}, \mathbf{x}_{m,n}) = \text{softmax}_{m,n}(\theta( \mathbf{x}_{i,j})^T \theta( \mathbf{x}_{m,n})),
\label{eq:fullattentionmap}
\end {equation}
where $\theta$ is a $1 \times 1$ convolution layer transforming the feature maps $\mathbf{x}$ to a new domain to calculate dot-product similarity \cite{cNonLocal} between every two pixels. Note that, following a common practice~\cite{cSPYGR}, we use the same $1 \times 1$ convolution weights for both query and key. Then, these similarities are used as the weights (Eq.~\eqref{eq:fullattentionapply}) to aggregate features of all pixels, producing an enhanced pixel representation $\mathbf{y}_{i,j}$ at position $(i,j)$.

\subsubsection {Formulation of the Axial Attention}
\label{sAxialAttentionFormulation}

From the above equations, we can see the computational complexity of the self-attention module is $O(H^2W^2)$, which requires large computation resources and prevents real-time applications, such as autopilot.
Several subsequent works~\cite{cCCNet, cAxialAttention} focused on reducing the computational complexity while maintaining high accuracy. In this work, we adopt axial-attention to perform spatial attention. In axial attention, the attention map is calculated for the column and row that cover the current pixel, reducing the computational complexity to $O(HW^2 +H^2W)$.

For convenience, we call the attention values calculated along the $Y$ axis ``\textit{column attention}'', and the attention values calculated along the $X$ axis ``\textit{row attention}''. Similar to Eq.~\ref{eq:fullattentionmap}, we define axial similarity functions by:
\begin {equation} 
\scriptsize
A_\text{col}( \mathbf{x}_{i,j},\mathbf{x}_{m,j})=\text{softmax}_{m}\left(\theta (\mathbf{x}_{i,j})^T \theta( \mathbf{x}_{m,j})\right) \ ,\ m\in [H]
\footnote{We use $i \in [ n ]$ to denote that $i$ is generated from $[ n ] = \{ {1,2,...,n} \}$.},
\label{eq:ColumnAttention}
\end {equation}
and
\begin {equation} 
\scriptsize
A_\text{row}( \mathbf{x}_{i,j}, \mathbf{x}_{i,n}) = \text{softmax}_{n}\left(\phi( \mathbf{x}_{i,j})^T \phi( \mathbf{x}_{i,n})\right) \ , \ n \in [W].
\label{eq:RowAttention}
\end {equation}
Note that we use different feature transformations ($\theta, \phi$) for the row and column attention calculations.

With the column and row attention maps $A_\text{col}$ and $A_\text{row}$, the final value weighted by the column and row attention maps can be represented by: 
\begin{equation}
\scriptsize
\mathbf{y}_{i,j} = \sum_{\forall n} \left( A_\text{row}(\mathbf{x}_{i,j}, \mathbf{x}_{i,n})  (\sum_{\forall m} A_\text{col}(\mathbf{x}_{i,j}, \mathbf{x}_{m,j}) g(\mathbf{x}_{m,n})) \right).
\label{eq:SumAcrossAxialAttention}
\end{equation}

\subsection {Channelized Axial Attention}
\label{sChannelizedAttention}

In order to address the feature conflicting issue of the existing dual attention designs, we propose a novel \textit{Channelized} Axial Attention (CAA), which seamlessly combines the advantages of spatial attention and channel attention.

As mentioned in the above sections, feature conflicts may be caused by the different interests of spatial and channel attentions. 
Ideally, channel attention should not ignore the regional features that are interesting to spatial attention. 
Conversely, spatial attention should consider channel relation as well.

Thus, we propose to compute channel attention within spatial attention. Specifically, we firstly break down spatial attention into more basic parts (Sect.~\ref{secBreakingDown}). Then, spatially varying channel attention is generated with $\bm{\alpha}_{i,j,m,n}$ and $\bm{\beta}_{i,j,n}$. In this way, channel attention is incorporated into spatial attention and spatial attention will not be ignored when small objects exist, seamlessly and effectively combining spatial and channel attention together.

\subsubsection{Breaking Down Axial Attention.}
\label{secBreakingDown}

For convenience, we firstly define two variables $\bm{\alpha}_{i,j,m,n}$ and $\bm{\beta}_{i,j,n}$ to represent the \textit{intermediate weighted features} as follows:
\begin{equation}
\scriptsize
\bm{\alpha}_{i,j,m,n}=A_\text{col}(\mathbf{x}_{i,j}, \mathbf{x}_{m,j})g(\mathbf{x}_{m,n})
\label{eq:ColAttentionApply}
\end{equation} 
and
\begin{equation}
\scriptsize
\bm{\beta}_{i,j,n}=A_\text{row}(\mathbf{x}_{i,j}, \mathbf{x}_{i,n})\sum_{\forall m}\bm{\alpha}_{i,j,m,n}. 
\label{eq:RowAttentionApply}
\end{equation}

Thus, Eq.~\eqref{eq:SumAcrossAxialAttention} can be rewritten as:
\begin{equation}
\scriptsize
\mathbf{y}_{i,j} =\sum_{\forall n} \bm{\beta}_{i,j,n} =\sum_{\forall n} A_\text{row}(\mathbf{x}_{i,j}, \mathbf{x}_{i,n})\left(\sum_{\forall m}\bm{\alpha}_{i,j,m,n}\right).
\label{eq:SumAcrossAxialAttentionSplit}
\end{equation}

Eqs.~\eqref{eq:ColAttentionApply},~\eqref{eq:RowAttentionApply} and~\eqref{eq:SumAcrossAxialAttentionSplit} show that the computation of the dot product is composed of two steps: 1) \textit{Reweighting}: re-weighting features on selected locations by column attention as in Eq.~\eqref{eq:ColAttentionApply} and row attention as in Eq.~\eqref{eq:RowAttentionApply}, and 
2) \textit{Summation}: summing the elements along row and column axes in Eq.~\eqref{eq:SumAcrossAxialAttentionSplit}.  Note that, this breakdown is also applicable to regular self-attention (see Table~\ref{tabCSA} and \textbf{Appendix}).

\subsubsection{Spatially Varying Channel Attention.}
With the intermediate results $\bm{\alpha}_{i,j,m,n}$ and $\bm{\beta}_{i,j,n}$ in Eqs.~\eqref{eq:ColAttentionApply} and~\eqref{eq:RowAttentionApply}, channel relation can be applied inside spatial attention, seamlessly combining them into one operation.
In addition, channel attention is now independently conducted on each column or row (on each pixel in regular self-attention) and provides spatial perspective for the channel relation modeling, resulting in our \textit{spatially varying channel attention}.
Enhanced with spatially varying channel attentions, now $C_\text{col}$ and $C_\text{row}$ are written as:
\begin{equation}
\scriptsize
C_\text{col}(\bm{\alpha}_{i,j,m,n}) = \text{Sigmod}\left( \text{ReLU}(\frac{ \sum_{\forall m,j}(\bm{\alpha}_{i,j,m,n})}{H \times W}\omega_{c1}) \omega_{c2} \right)\bm{\alpha}_{i,j,m,n},
\label{eq:ChannelAttentionColumn}
\end{equation}
and
\begin{equation}
\scriptsize
C_\text{row}(\bm{\beta}_{i,j,n}) = \text{Sigmod}\left( \text{ReLU}(\frac{ \sum_{\forall i,n}(\bm{\beta}_{i,j,n})}{H \times W} \omega_{r1}) \omega_{r2} \right)\bm{\beta}_{i,j,n},
\label{eq:ChannelAttentionRow}
\end{equation}
where $\text{Sigmod} (\cdot) $ is the learned channel attention, and $\omega_{c1}$, $\omega_{c2}$, $\omega_{r1}$ and $\omega_{r2}$ are the learned relationships between different channels according to $\bm{\alpha}_{i,j,m,n}$ and $\bm{\beta}_{i,j,n}$.

Thus, instead of directly using $\bm{\alpha}_{i,j,m,n}$ and $\bm{\beta}_{i,j,n}$ as in Eq.~\eqref{eq:SumAcrossAxialAttentionSplit}, 
for each column and row, we obtain the channelized axial attention features, 
where the intermediate results $\bm{\alpha}_{i,j,m,n}$ and $\bm{\beta}_{i,j,n}$ are weighted by the spatially varying channel attention defined in  Eqs.~\eqref{eq:ChannelAttentionColumn} and~\eqref{eq:ChannelAttentionRow} as: 
\begin{equation}
\scriptsize
\mathbf{y}_{i,j} = \sum_{\forall n} C_\text{row}\left(A_\text{row}(\bm{x}_{i,j}, \bm{x}_{i,n})(\sum_{\forall m} C_\text{col}(\bm{\alpha}_{i,j,m,n}) )\right). 
\label{eq:CAAFinal}
\end{equation}

Note that the spatially varying channel attention keeps a $W$ dimension after averaging $H\times W$ during the channel attention (Fig.~\ref{figOverall}). 
Now each row has its own channel attention thanks to the breaking down of spatial axial attention.



\subsubsection{Going Deeper in Channel Attention.}
\label{sGoingDeeper}

Similar to the work in~\cite{cSENet}, we use two fully connected layers, followed by ReLU and sigmoid activations respectively, to first reduce the channel number and then increase it to the original channel number.

To further boost performance, we explore the design of more powerful channel attention modules for our channelization since our attention module keeps the spatial dimension, and thus contains more information than a regular SE module ($1\times 1\times C\times W or H$ vs $1\times 1\times C $, see Fig.~\ref{figOverall}).

We experimented with increased depth and/or width of hidden layers to enhance the capacity of spatial varying channel attention. 
We find that deeper hidden layers allow channel attention to find a better relationship between channels for our spatially varying channel attention.
Moreover, increasing layer width is not as effective as adding layer depth (see Table~\ref{table:cdp}).

Furthermore, in spatial domain, each channel of a pixel contains unique information that can lead to a unique semantic representation. 
We find that using Leaky ReLU~\cite{cLeakyrelu} is more effective than ReLU in preventing the loss of information along deeper activations~\cite{cMobileNetV2}. 
Apparently, this replacement only works in spatially varying channel attention.

\subsubsection{Grouped Vectorization.} \label {sec:secGroupVect}

Computing spatial attention row by row and column by column can save computation but it is still too slow (about 2.5 times slower on a single V100 with feature map size = $33\times33$) even with parallelization. 
Full vectorization can achieve a very high speed but it has a high requirement on GPU memory (about 2 times larger GPU memory usage than no vectorization on a single V100 with feature map size = $33\times33$) for storing the intermediate partial axial attention results $\bm{\alpha}$ (which has a dimension of $H \times H \times W \times C$) and $\bm{\beta}$ (which has a dimension of $W\times H \times W \times C$) in Eqs.~\eqref{eq:ColAttentionApply} and~\eqref{eq:RowAttentionApply}. 
To enjoy the high speed benefit of vectorization with limited GPU memory usage, in our implementation we propose \textit{grouped} vectorization to dynamically batch rows and columns into multiple groups, and then perform vectorization for each group individually. 
The technical details and impact of our group vectorization are detailed in the \textbf{Appendix}.


\section{Experiments}
\label{sExperements}

To demonstrate the effectiveness for accuracy of the proposed CAA, comprehensive experimental results are compared with the state-of-the-art methods on three benchmark datasets, \textit{i.e.}, PASCAL Context~\cite{cPascalVOC}, COCO-Stuff~\cite{cCocoStuff} and Cityscapes~\cite{cCityScapes}. 

Using similar settings as in other existing works, we measure the segmentation accuracy using mean intersection over union (mIOU). 
Moreover, to demonstrate the efficiency of our CAA, we also compare the floating point operations per second (FLOPS) of different approaches. 
Experimental results show that our CAA outperforms the state-of-the-art methods on all tested datasets. 
Due to page limitations, we focus on ResNet-101 (with naive upsampling) results in the main paper for the fairest comparison. Results obtained with EfficientNet or results on other extra datasets are presented in the \textbf{Appendix}.

\subsection{Implementation Details}

\subsubsection{Backbone} 
Our network is built on ResNet-101~\cite{cResnet} pre-trained on ImageNet. 
The original ResNet results in a feature map of ${1}/{32}$ of the input size. 
Following other works~\cite{cDeepLabV3Plus, cEMANet}, we apply dilated convolution at the output stride = 16 for ablation experiments if not specified. We conduct experiments with the output stride = 8 to compare with the state-of-the-art methods.

\subsubsection{Naive Upsampling} Unless otherwise specified, we directly bi-linearly upsampled the logits to the input size without refining using any low-level and high resolution features.

\subsubsection{Training Settings} 
We employ stochastic gradient descent (SGD) for optimization, where the polynomial decay learning rate policy $ (1 - \frac{iter}{max iter})^{0.9} $ is applied with an initial learning rate = 0.01. 
We use synchronized batch normalization with batch size = 16 (8 for Cityscapes) during training.
For data augmentation, we only apply the most basic data augmentation strategies as in~\cite{cDeepLabV3Plus}, including random flip, random scale and random crop.

\subsection{Results on PASCAL Context}
\label{expAblation}

The PASCAL Context~\cite{cPascalContext} dataset has 59 classes with 4,998 images for training and 5,105 images for testing. 
We train our CAA on the training set for 40k iterations.
In the following, we first present a series of ablative experiments to show the effectiveness of our method. 
Then, quantitative and qualitative comparisons with other state-of-the-art methods are presented.

Note that, in ablation studies below, we report mean result with standard deviation (numbers in parentheses) calculated with 5 repeated experiments. (See \textit{Deterministic} in the \textbf{Appendix} for technical details).
\begin{table}[t]
	\centering
	\small
	\begin{tabular}{c|c|c|c} 
		\toprule[1pt]
		\multicolumn{1}{c|}{Layer Counts} &\multicolumn{1}{c|}{\# of Channels} & mIOU (\%) & FLOPs \\
		\midrule[0.5pt]
		\midrule[0.5pt]
		-& -& 50.27($\pm$0.2)&  68.7G\\
		\midrule
		1&128& 50.75($\pm$0.2)& +0.00024G\\
		3&128& 50.85($\pm$0.2) & +0.00027G\\
		5&128& \textbf{51.06($\pm$0.2)} & +0.00030G \\
		7&128& 50.40($\pm$0.3)  & +0.00043G \\
		\midrule
		5&64& 50.12($\pm$0.2)& +0.00015G\\
		5&256 & 50.35($\pm$0.4) & +0.00098G \\
		\bottomrule[1pt]
	\end{tabular}
	\caption{ Results without using channelization (Row 1) and using channelization with different layer counts and channel numbers. Numbers in parentheses indicate standard deviations (see Sect.~\ref{expAblation}).}
	\label{table:cdp}
\end{table}

\subsubsection{Effectiveness of the Proposed Channelization}
We first report the impact of adding channelized axial attention and with different depth and width in \tablename{~\ref{table:cdp}}, where `-' for the baseline result indicates no channelization is performed.

\begin{table}[t]
    \centering
    \small
    \begin{tabular}{c|c|c}
        \toprule[1pt]
        Axial Attention& + SE & + Our Channelization \\
        \midrule[0.5pt]
		\midrule[0.5pt]
        50.27($\pm$0.2) & 50.37($\pm$0.2)& 51.06($\pm$0.2) \\
        \bottomrule[1pt]
    \end{tabular}
     \caption{ Result comparison between axial attention, axial attention + SE and channelized axial attention.  }
    \label{table:sedesign}
\end{table}

As can be seen from this table, our proposed channelization improves the mIOU over the baseline regardless of the layer counts and the number of channels used. 
In particular, the best performance is achieved when the Layer Counts = 5 and the number of Channels = 128. 

We also compare our model with the sequential design of ``Axial Attention + SE'', as shown in \tablename{~\ref{table:sedesign}}. 
We find the sequential design brings only marginal contributions to performance, showing that our proposed channelization method can combine the advantages of both spatial attention and channel attention more effectively. 
In \tablename{~\ref{tabBackbones}}, results obtained with \textit{other backbones} are provided to demonstrate the effectiveness and robustness of CAA.

\subsubsection{Channelized Self-Attention}
In this section, we conduct experiments on the PASCAL Context by applying channelization to the original self-attention. 
We report its single-scale performance in~\tablename{ \ref{tabCSA}} with ResNet-101. 
Our channelized method can also further improve the performance of self-attention by 0.67\% (51.09\% vs 50.42\%).

\begin{table}[t]
	\centering
	\small
	\begin{tabular}{l|c|c|c} 
		\toprule
		Attention Base                    & Eval OS & Channelized     & mIOU (\%)   \\
		\midrule[0.5pt]
		\midrule[0.5pt]
		\multirow{2}{*}{Axial Attention  }& 16       &             & 50.27  \\
		& 16  	 & \checkmark  & 51.06  \\
		\midrule
		\multirow{2}{*}{  Self Attention} & 16  	 &             & 50.42  \\
		& 16  	 & \checkmark  & 51.09  \\
		\bottomrule[0.5pt]
	\end{tabular}
	\caption{Ablation study of applying our Channelized Attention on self-attention with ResNet-101. \textbf{Eval OS}: Output strides~\cite{cDeepLabV3Plus} during evaluation.}
	\label{tabCSA}
\end{table}

\subsubsection{Impact of the Testing Strategies} 
We compare the performance and computation cost of our proposed model against the baseline and DANet~\cite{cDualAttention} with different testing strategies in \tablename{~\ref{table:teststrategies}}.
Using the same settings as in other works~\cite{cDualAttention}, we add multi-scale, left-right flip and auxiliary loss during inference. 
The accuracies of CAA are further boosted with output stride = 8 since the channel attention can learn and optimize three times more pixels.

\begin{table}[t]
	\centering
	\small
	\begin{tabular}{c|c|c|c|c|c} 
		\toprule[1pt]
		Methods&OS& MF &Aux& mIOU (\%) & FLOPs\\ 
		\midrule[0.5pt]
		\midrule[0.5pt]
		ResNet& 16 &   & - & - & 59.85G \\
		-101& 8 &   & - & - & 190.70G  \\
		\midrule
		DANet& 8 &   & &  & +101.25G\\ 
		     & 8 & \checkmark  & \checkmark &52.60 & -   \\ 
		\midrule
		Axial     & 16    &               &       & 50.27($\pm$0.2)&  +8.85G\\ 
		Attention & 16    & \checkmark    &       & 52.01($\pm$0.2) & -\\ 
		          & 8     &               &       & 51.24($\pm$0.2) & +34.33G\\ 
		          & 8     & \checkmark    &       &52.51($\pm$0.2)  & -   \\ 
		\midrule
		Our       & 16    &               &     & 51.06($\pm$0.2)&  +8.85G\\
		CAA       & 16    & \checkmark    &   &53.09($\pm$0.3)      & -  \\
		          & 8     &               &       & 52.73($\pm$0.1) & +34.33G\\ 
		          & 8     & \checkmark    &       & 54.05($\pm$0.1) & -   \\ 
		\midrule
		Our   & 16    &               & \checkmark & 51.80($\pm$0.2)&  +8.85G\\
		 CAA         & 16 &\checkmark& \checkmark & 53.52($\pm$0.2)& - \\
		 + & 8 &          & \checkmark& 53.48($\pm$0.3)   & +34.33G\\
		 Aux loss  & 8 & \checkmark   &\checkmark &54.65($\pm$0.4)      & - \\
		\bottomrule[1pt]
	\end{tabular}
	\caption{Comparison results with different testing strategies. 
		\textbf{OS:} Output stride in training and inference. \textbf{MF:} Apply multi-scale and left-right flipping during inference. 
		\textbf{Aux:} Add auxiliary loss during training.
		``$\bm{+}$'' refers to the extra FLOPS over the baseline FLOPS of ResNet-101. }
	\label{table:teststrategies}
\end{table}

\begin{table}[ht]
	\centering
	\small
	\begin{tabular}{l |c|c|c|c}
		\toprule
		Backbone 				   										& OS  	&AA   & C   & mIOU (\%)     \\
		\midrule[0.5pt]
		\midrule[0.5pt]
		ResNet-50
		& 16       	&\checkmark		    &              	& 49.73     \\
		~\cite{cResnet}& 16  	   	&\checkmark		    & \checkmark  	& \textbf{50.23}     \\
		\midrule
		Xception65
		& 16       	&\checkmark		    &              	& 52.42     \\
		~\cite{cXception}& 16  	   	&\checkmark		    & \checkmark  	& \textbf{52.65}     \\
		\midrule
		EfficientNetB7
		& 16       	&\checkmark		    &              	& 57.24     \\
		~\cite{cEfficientNet}& 16  	   	&\checkmark		    & \checkmark  	& \textbf{57.93}     \\
		& 8  	   	&\checkmark		    & \checkmark  	& \textbf{58.40}     \\         
		
		\bottomrule[0.5pt]
		
	\end{tabular}
	\caption{Ablation study of other backbones.
		All results are obtained in single scale without flipping.
		\textbf{OS}: Output strides during evaluation.
		\textbf{AA}: Axial Attention.
		\textbf{C}: Channelized.}
	\label{tabBackbones}
	
\end{table}

\begin{table}[ht]
	\centering
	\small
	\begin{tabular}{l|c|c}
		\toprule[1pt]
		\rule{0pt}{2ex} Methods & Backbone & mIOU (\%) \\
		\midrule[0.5pt]
		\midrule[0.5pt]
		ENCNet~\cite{cENCNet} & ResNet-101 & 51.7 \\
		ANNet~\cite{cANNN} & ResNet-101 & 52.8 \\
		EMANet~\cite{cEMANet} & ResNet-101 & 53.1 \\
		SPYGR~\cite{cSPYGR} & ResNet-101 & 52.8 \\
		CPN~\cite{cCPN} & ResNet-101 & 53.9 \\
		CFNet~\cite{cCFNet} & ResNet-101 & 54.0 \\
		\midrule[0.5pt]
		DANet~\cite{cDualAttention} & ResNet-101 & 52.6\\
		\midrule[0.5pt]
		Our CAA (OS = 16) & ResNet-101 & 53.7 \\
		\textbf{Our CAA (OS =\enspace8)} & ResNet-101 & \textbf{55.0} \\
		\bottomrule[1pt]
	\end{tabular}
	\caption{Comparisons with other state-of-the-art approaches on the PASCAL Context test set. For a fair comparison, all compared methods used ResNet-101 and naive upsampling.}
	\label{tabPascalContextSOTA}
\end{table}

\subsubsection{Comparison with the State-of-the-art}
\label{lab:PascalCtxSOTA}

Finally,  in \tablename{~\ref{tabPascalContextSOTA}}, we compare our approach with the state-of-the-art approaches. 
Like other similar works, we apply multi-scale and left-right flip during inference. 
For a fair comparison, we only compare with the methods that use ResNet-101 and naive upsampling in the main paper. More results using alternative backbones are included in \tablename{~\ref{tabBackbones}}.

As shown in this table, our proposed CAA outperforms all listed state-of-the-art models that are trained with an output stride = 8. 
Our CAA also performs better than EMANet and SPYGR that are trained with output stride = 16.  
Note that, in this and the following tables, we report the best results of our approach obtained in experiments.

\begin{figure}[t]
	\centering
	\includegraphics[width=0.98\linewidth]{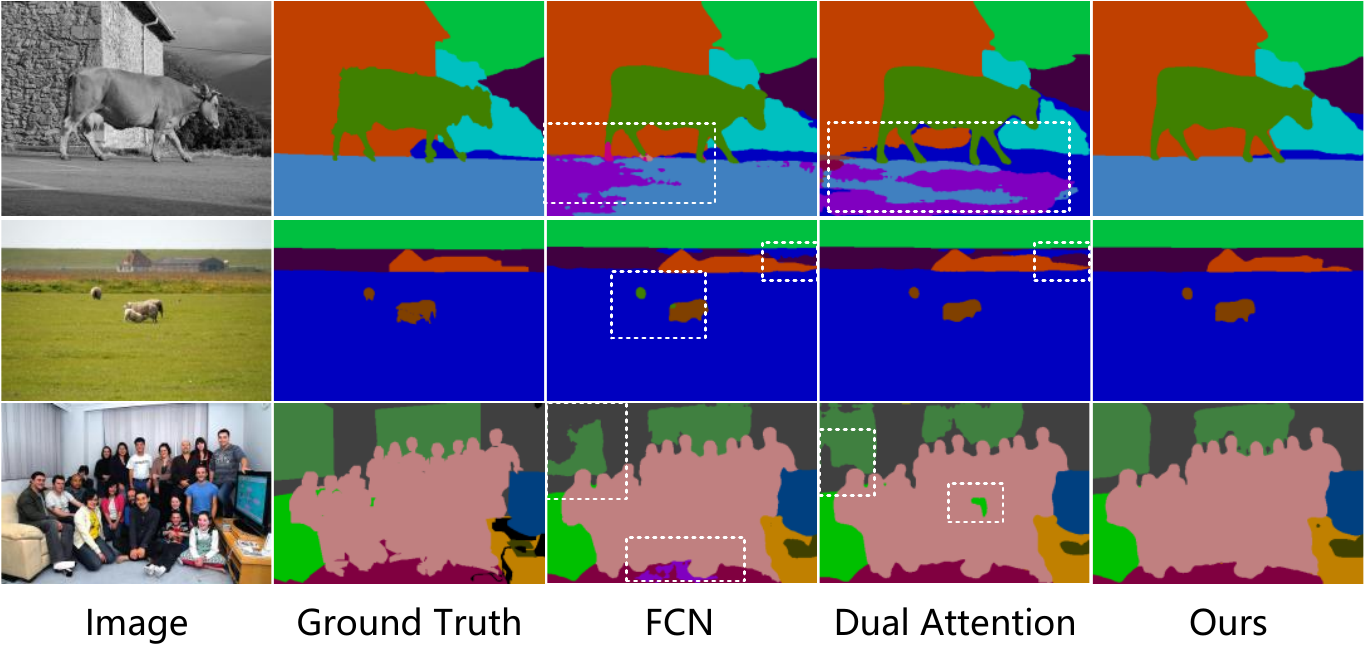}
	\caption{Examples of the segmentation results obtained on the PASCAL Context dataset using FCN, DANet and CAA. }
	\label{fvis0}
\end{figure}

In Fig.~\ref{fvis0}, we show some results obtained by our CAA model, FCN and Dual attention.
Our model is able to handle previous failure cases better, especially when a class A covering a smaller region is surrounded by another class B covering a much larger region (see the boxed regions).

\subsection{Results on the COCO-Stuff 10K}

Following the other works~\cite{cDualAttention}, we evaluate our CAA on COCO-Stuff 10K dataset~\cite{cCocoStuff}, which contains 9,000 training images and 1,000 testing images with 172 classes. Our model is trained for 40k iterations.
As shown in \tablename{~\ref{tabCocostuffSOTA}}, our proposed CAA outperforms all other state-of-the-art approaches by a large margin of $1.3\%$, demonstrating that our model can better handle complex images with a large number of classes.

\begin{table}[t]
	\centering
	\small
	\begin{tabular}{l|c|c}
		\toprule[1pt]
		\rule{0pt}{2ex} Methods & Backbone & mIOU (\%) \\
		\midrule[0.5pt]
		\midrule[0.5pt]
		DSSPN~\cite{cDSSPN} & ResNet-101 & 38.9 \\
		SVCNet~\cite{cSVCNet} & ResNet-101 & 39.6 \\
		EMANet~\cite{cEMANet} & ResNet-101 & 39.9 \\
		SPYGR~\cite{cSPYGR} & ResNet-101 & 39.9 \\
		OCR~\cite{cOCR} & ResNet-101 & 39.5 \\
		\midrule[0.5pt]
		DANet~\cite{cDualAttention} & ResNet-101 & 39.7\\
		\midrule[0.5pt]
		\textbf{Our CAA} & ResNet-101 & \textbf{41.2} \\
		\bottomrule[0.5pt]
	\end{tabular}
	\caption{Comparisons with other state-of-the-art approaches on the COCO-Stuff 10K test set. For a fair comparison, all compared methods use ResNet-101 and naive upsampling.}
	\label{tabCocostuffSOTA}
\end{table}

\begin{table}[!t]
	\centering
	\small
	\begin{tabular}{l|c|c}
		\toprule[1pt]
		\rule{0pt}{2ex} Methods & Backbone & mIOU (\%) \\
		\midrule[0.5pt]
		\midrule[0.5pt]
		CFNet~\cite{cCFNet} & ResNet-101 & 79.6 \\
		ANNet~\cite{cANNN} & ResNet-101 & 81.3 \\
		CCNet~\cite{cCCNet} & ResNet-101 & 81.4 \\
		CPN~\cite{cCPN} & ResNet-101 & 81.3 \\
		SPYGR~\cite{cSPYGR} & ResNet-101 & 81.6 \\
		OCR~\cite{cOCR} & ResNet-101 & 81.8 \\
		\midrule[0.5pt]
		DANet~\cite{cDualAttention} & ResNet-101 & 81.5\\
		\midrule[0.5pt]
		\textbf{Our CAA} & ResNet-101 & \textbf{82.6} \\
		\bottomrule[1pt]
	\end{tabular}
	\caption{Comparisons with other state-of-the-art approaches on the Cityscapes Test set. For a fair comparison, all compared methods use ResNet-101 and naive upsampling.}
	\label{tabCityscapesSOTA}
\end{table}

\subsection{Results on the Cityscapes}

The Cityscapes dataset~\cite{cCityScapes} has 19 classes. Its \textit{fine} set contains high quality pixel-level annotations of 2,975, 500 and 1,525 images in the training, validation, and test sets, respectively.
Following previous works~\cite{cDualAttention}, we use only the \textit{fine} set with a crop size of 769$\times$769 during training, and our training iteration is set to 80k.
We report our results on the \textit{test} set in \tablename{~\ref{tabCityscapesSOTA}}. Results show our CAA is also working well on high-resolution images.

\section{Conclusion}
\label {secConclusion}

In this paper, we have proposed a novel and effective Channelized Axial Attention, effectively combining the advantages of the popular spatial-attention and channel attention. 
Specifically, we first break down spatial attention into two steps and insert channel attention in between, enabling different spatial positions to have their own channel attentions.
Experiments on the three popular benchmark datasets have demonstrated the effectiveness of our proposed channelized axial attention.

\section{Acknowledgments}
We thank TensorFlow Research Cloud (TFRC)
for TPUs.

\clearpage

\appendix

\section{Appendix}

\subsection{Extra Experiments}
\subsubsection{Stronger Backbone in PASCAL Context}

As mentioned in the main paper, our CAA outperforms the SOTA methods~\cite{cCFNet, cEMANet} with the same settings (ResNet-101 + naive upsampling). 
Furthermore, we show our proposed CAA is suitable for different backbones.

In this section, we report our CAA's performance with EfficientNet~\cite{cEfficientNet} in \tablename{~\ref{tabPascalContextFree}}. 
Note that, this is not a fair comparison, since the listed methods were not trained under the same settings, or using the same backbone. 
The results show that our method can  outperform the state-of-the-art Transformer~\cite{cViT} based hybrid models such as SETR~\cite{cSETR} and DPT~\cite{cDPT} with the CNN backbone EfficientNet-B7. 
The \textit{simple decoder} merges the low level features from output stride = 4, during the final upsampling (see \cite{cDeepLabV3Plus} for details).

\begin{table}[t]
	\centering
	\small
	\begin{tabular}{l|c}
		\toprule[1pt]
		\rule{0pt}{2ex} Methods & mIOU (\%) \\
		\midrule[0.5pt]
		\midrule[0.5pt]
		CTNet~\cite{cTNet} + JPU & 55.5 \\
		SETR-MLA~\cite{cSETR} & 55.83 \\
		HRNetV2 + OCR ~\cite{cSVCNet} & 56.2\\
		ResNeSt-269~\cite{cResnest} + DeepLab V3+ & 58.9 \\
		HRNetV2 + OCR + RMI& 59.6 \\
		DPT-Hybrid~\cite{cDPT} & 60.46 \\
		\midrule[0.5pt]
		Our CAA (EfficientNet-B7, w/o decoder) & 60.12 \\
		\textbf{Our CAA (EfficientNet-B7 + simple decoder) } & \textbf{60.50}\\
		\bottomrule[1pt]
	\end{tabular}
	\caption{Result comparison with the state-of-the-art approaches on the PASCAL Context test set for multi-scale prediction. Note that, the listed methods were not trained under the same settings, or using same backbone.}
	\label{tabPascalContextFree}
\end{table}

\subsubsection{Stronger Backbone in COCOStuff-10K}
We also report our CAA's results using Efficientnet-b7~\cite{cEfficientNet} as the backbone in \tablename{~\ref{tabCocostuffFree}}. 

\subsubsection{Results in COCOStuff-164k}

The recent method Segformer~\cite{cSegFormer} used COCOStuff-164k (164,000 images), \textit{i.e.}, the full set of COCOStuff-10k to validate its performance for the first time. Since Segformer is a strong backbone, in this section, we also use EfficientNet-B5 + CAA to verify the robustness of our CAA on COCOStuff-164k. 
Table~\ref{tabCocostuff164KFree} shows that our method outperforms the recent strong baselines Segformer and SETR~\cite{cSETR} by a large margin, indicating our CAA keeps the superior performance with large training data.

\begin{table}[t]
	\centering
	\small
	\begin{tabular}{l|c}
		\toprule[1pt]
		\rule{0pt}{2ex} Methods & mIOU (\%) \\
		\midrule[0.5pt]
		\midrule[0.5pt]
		HRNetV2 + OCR ~\cite{cSVCNet} & 40.5\\
		DRAN & 41.2 \\
		HRNetV2 + OCR + RMI & 45.2 \\
		\midrule[0.5pt]
		\textbf{Our CAA (EfficientNet-B7)} & \textbf{45.4}\\
		\bottomrule[1pt]
	\end{tabular}
	\caption{Result comparison with the state-of-the-art approaches on the COCO-Stuff-10K test set for multi-scale prediction. Note that, the listed methods were not trained under the same settings, or using same backbone. }
	\label{tabCocostuffFree}
\end{table}

\begin{table}[t]
	\centering
	\small

	\begin{tabular}{l|c}
		\toprule[1pt]
		\rule{0pt}{2ex} Methods & mIOU (\%) \\
		\midrule[0.5pt]
		\midrule[0.5pt]
		ResNet-50 + DeepLabV3+ ~\cite{cDeepLabV3Plus} & 38.4\\
		HRNetV2 + OCR & 42.3 \\
		SETR~\cite{cSETR} & 45.8\\
	    Segformer-B5~\cite{cSegFormer} & 46.7 \\
		\midrule[0.5pt]
		\textbf{Our CAA (EfficientNet-B5)} & \textbf{47.30}\\
		\bottomrule[1pt]
	\end{tabular}
	\caption{Result comparison with the state-of-the-art approaches on the COCO-Stuff-164K test set for multi-scale prediction. Note that, the listed methods were not trained under same settings, or using same backbone. \textit{Methods} other than CAA and Segformer are reproduced in Segformer paper. }
	\label{tabCocostuff164KFree}
\end{table}

\subsection{Extra Visualizations}

\begin{figure}[t]
	\centering
	\includegraphics[width=\linewidth]{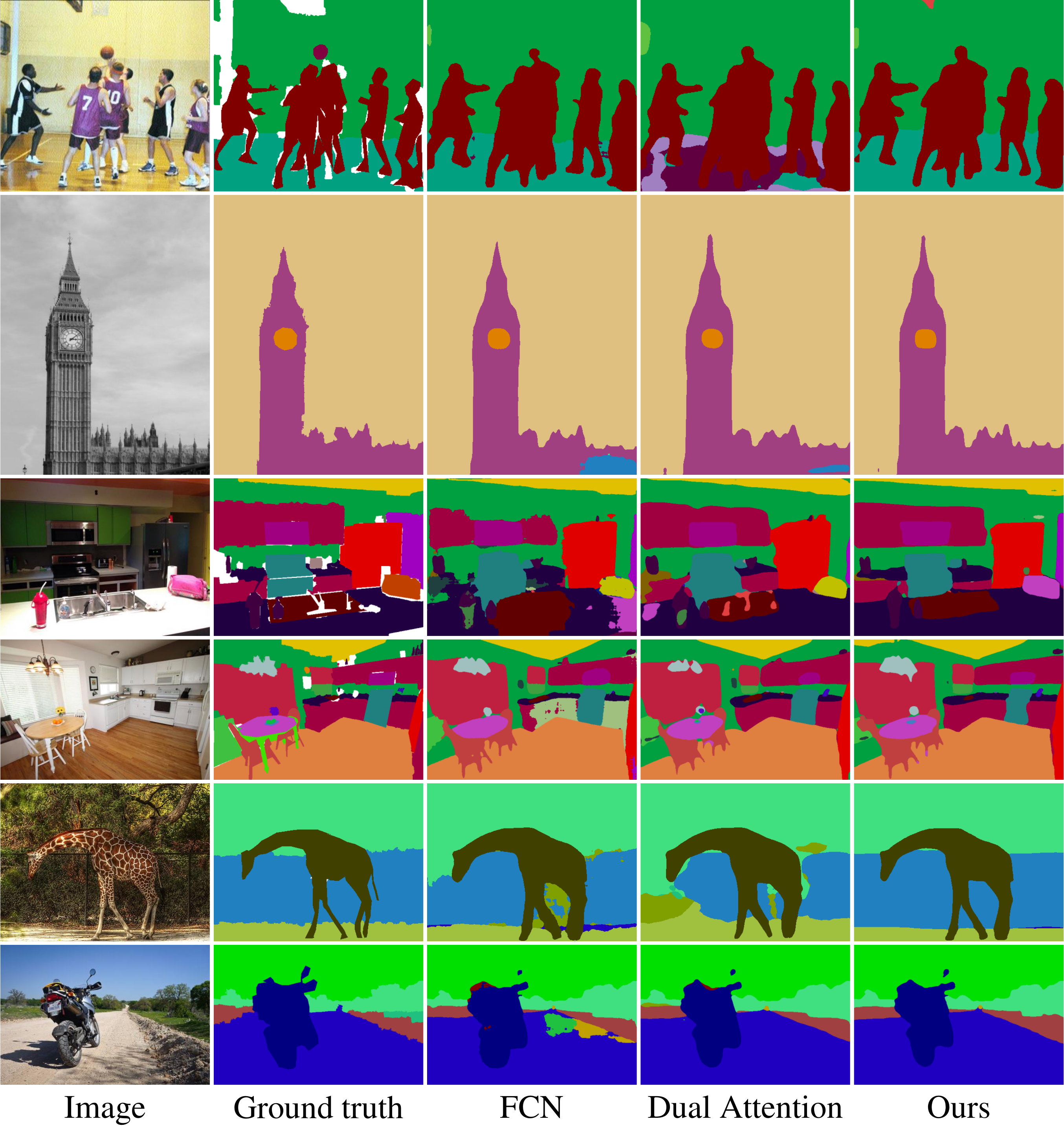}
	\caption{Examples of the results obtained on the COCO-Stuff 10K dataset with our proposed CAA in comparison to the results obtained with FCN, DANet and the ground truth.} 
	\label{fviscocostuff}
\end{figure}

\begin{figure}[t]
    \centering
    \includegraphics[width=\linewidth]{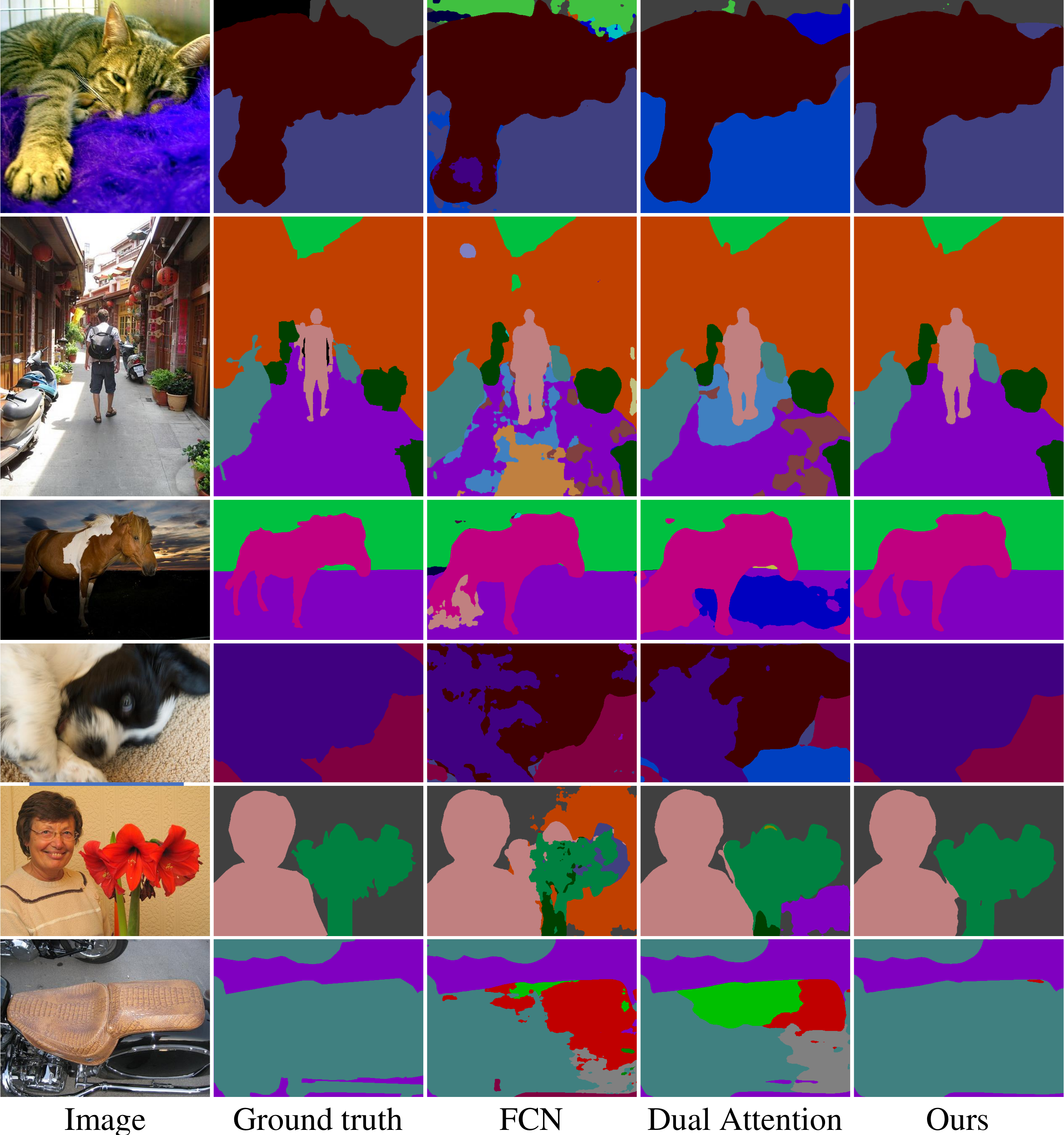}
    \caption{Examples of the results obtained on the PASCAL Context dataset with our proposed CAA in comparison to the results obtained with FCN, DANet and the ground truth.}
    \label{fig:extrapascalctx}
\end{figure}

\begin{figure*}[t]
    \centering
    \includegraphics[width=0.9\linewidth]{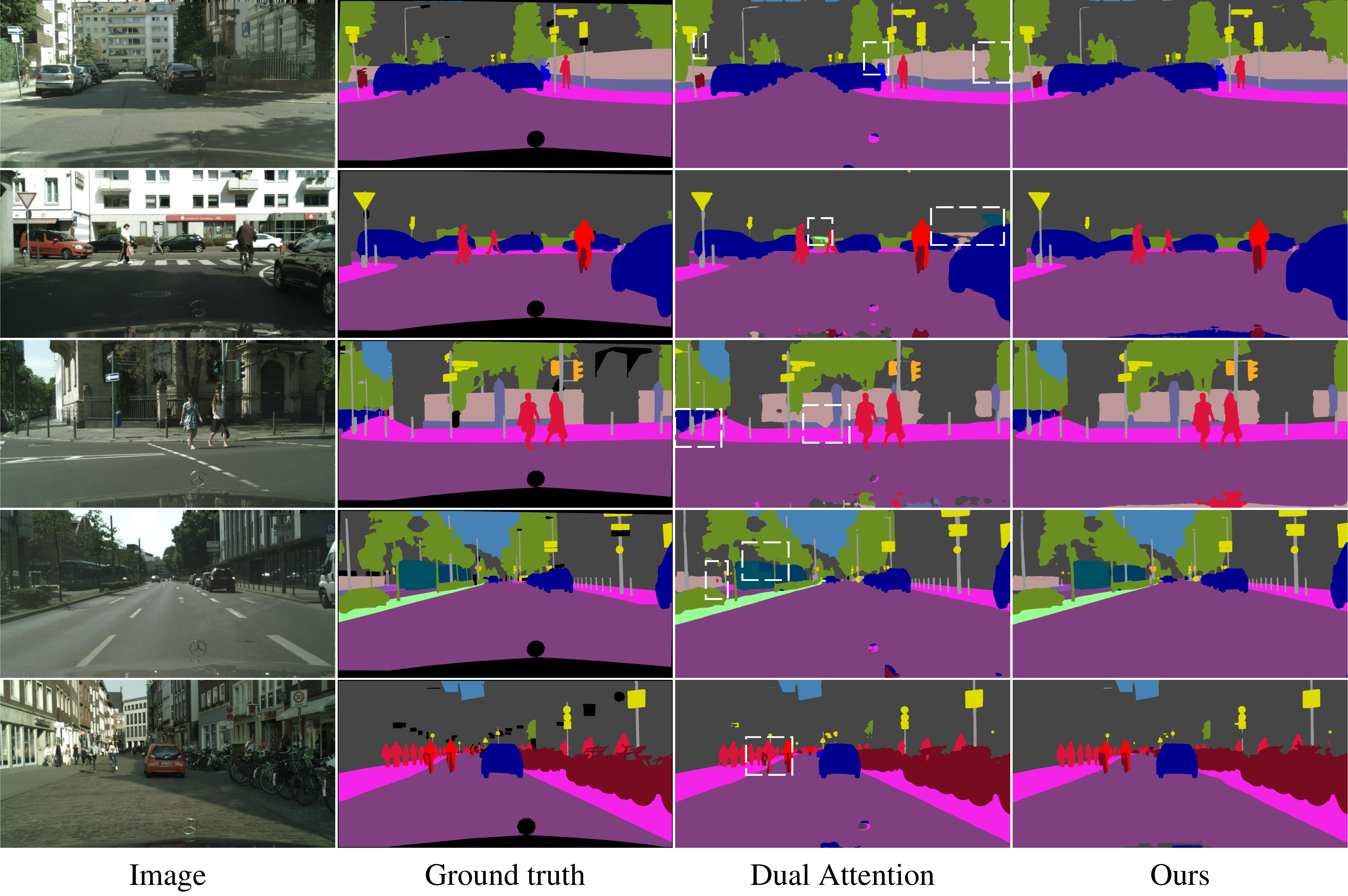}
    \caption{Extra examples of the segmentation results obtained on the Cityscapes validation set~\cite{cCityScapes} with our proposed CAA in comparison to the results obtained with DANet~\cite{cDualAttention} and the ground truth.}
    \label{fig:viscityscapes}
\end{figure*}

\subsubsection{COCOStuff-10k}

Fig.~\ref{fviscocostuff} shows some examples of the segmentation results obtained on the COCO-Stuff 10K with our proposed CAA in comparison to the results of FCNs~\cite{cFCN}, DANet~\cite{cDualAttention}, and the ground truth (output stride = 8, ResNet-101). 
As it can be seen, our CAA can segment common objects such as building, human, or sea very well. 

\subsubsection{Extra PASCAL Context} In the main paper, due to the page limit, only 3 images from PASCAL Context are presented. 
In this section, we show more examples of the segmentation results obtained on the PASCAL Context in Fig.~\ref{fig:extrapascalctx}. Results show that the failure cases in FCN and DANet are segmented much better by our CAA, especially hard cases (see the 2nd row).

\subsubsection{Cityscapes} In Fig.~\ref{fig:viscityscapes}, we compare the segmentation results obtained on Cityscapes validation set with DANet and our CAA. 
Key areas of difference are highlighted with white boxes. 
Results show that many errors produced by DANet no longer exist in our CAA results.

\subsection{Group Vectorization}

Algorithm~\ref{alg:groupvect} presents the pseudo code of implementing the proposed grouped vectorization. 

\begin{algorithm}[t]
    \small
    \setlength{\abovecaptionskip}{0.1cm}
    \setlength{\belowcaptionskip}{0.1cm}

    \caption{Our proposed grouped vectorization algorithm}

    \begin{algorithmic}[1]
        \Require $G$: Group Number, $A$: Attention Map $[N, H, H, W]$, $X$: Feature Map $[N, C, H, W]$
        
        \State $padding \gets H\:\%\:G$
        \State $A\gets$ Transpose $A$ into $ [H, N, H, W] $
        \State $H^{+} \gets H + padding$
        \State $A\gets$ padding zero to  $A$ into $[H^{+}, N, H, W]$
        \State $A\gets$ Reshape $A$ into $[G, H^{+}\://\:G, N, H, W]$
        \For{$g\in G$}
            \State $Y_{g}\gets$ Channelization $(X, A_{g}),\:Y_{g}\in[H^{+}\://\:G, N, C, W]$
        \EndFor
        \State $Y\gets$ Concat$(Y_{0,1,...G}),\:Y\in[G, H^{+}\://\:G, N, C, W]$
        \State $Y\gets$ Reshape  $Y$ into $[H^{+}, N, C, W]$
        \State $Y\gets$ Remove padding from $Y$ into $[H, N, C, W]$
        \State $Y\gets$ Transpose $Y$ into $[N, C, H, W]$
        
        \Return Y
    \end{algorithmic}
    \label{alg:groupvect}
\end{algorithm}

\subsubsection{Effectiveness of Our Grouped Vectorization}

In our main paper, we introduced the grouped vectorization to split tensors into multiple groups so as to reduce the GPU memory usage when preforming channel attention inside spatial attention.
As we use more groups in group vectorization, the proportionally less GPU memory is needed for the computation. 
However, longer running time is required.  
In this section, we conduct experiments to show the variation of the inference time (seconds/image) when different numbers of groups are used. 

\begin{figure}[t]
	\centering
	\includegraphics[width=3in]{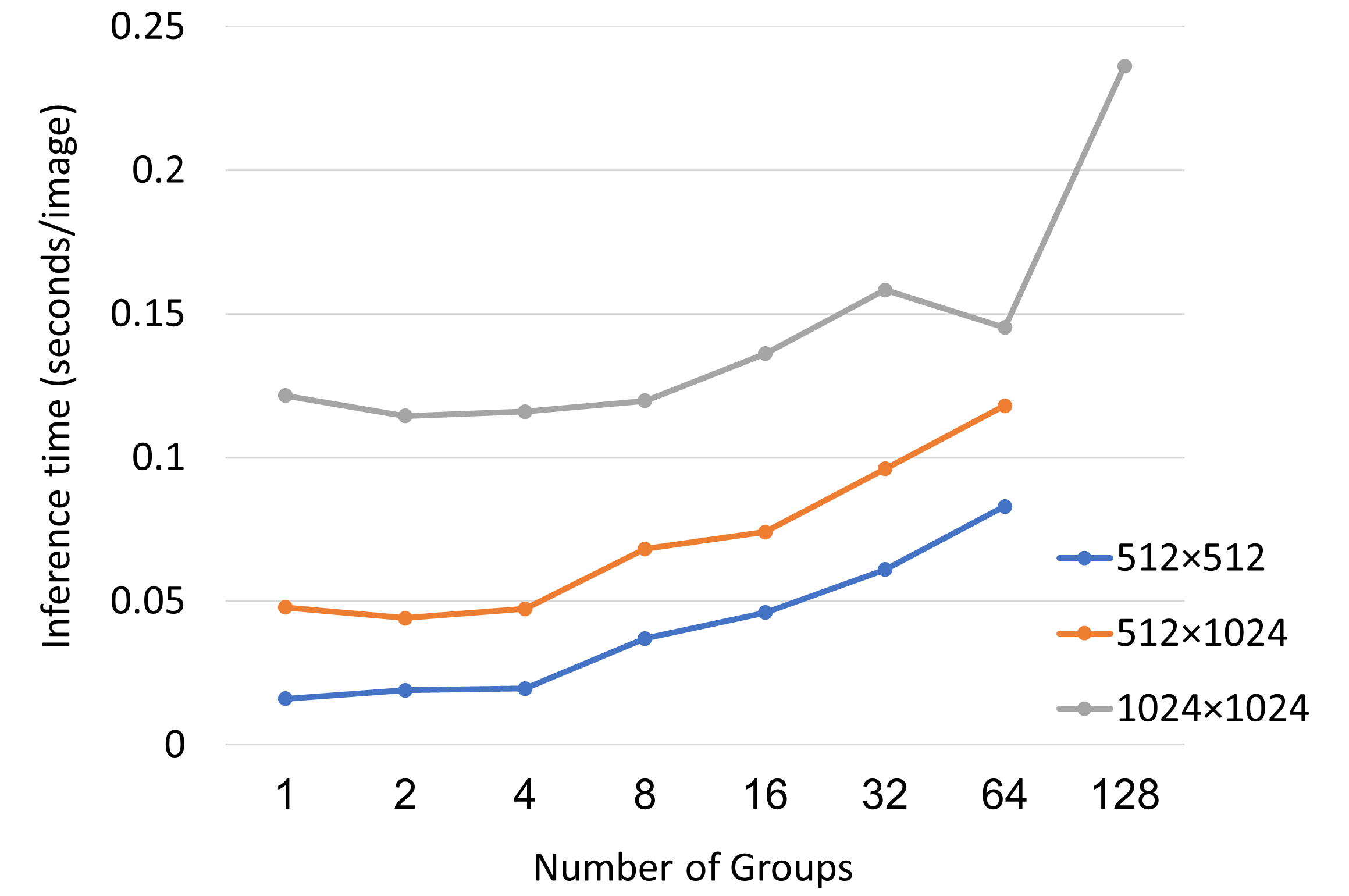}
	\caption{Inference time (seconds/image) when applying different numbers of groups in grouped vectorization.}
	\label{figSpeedVSGroups}
\end{figure}

Fig.~\ref{figSpeedVSGroups} shows the results of three different input resolutions. 
As shown in this graph, when splitting the vectorization into smaller numbers of groups, \textit{e.g.}, 2 or 4, our grouped vectorization can achieve similar inference speed using one half or one quarter of the original spatial complexity. 
For example, separating into 4 groups has similar inference speed with no separation (1 group).

\subsection{Extra Technical Details}

\subsubsection{Deterministic}
In our paper, all experiments are conducted with enabled deterministic and fixed seed value to reduce the effect of randomness. 
However, in current deep learning frameworks (\textit{e.g.}, PyTorch or TensorFlow), completely reproducible results are not guaranteed since not all the operations have a deterministic implementation. 
To show the robustness of our proposed method, in the ablation studies of the main paper, we repeat each experiment 5 times and report their mean results and standard deviation. 
To learn more about deterministic, please check ``https://pytorch.org/docs/stable/notes/randomness.html" and ``https://github.com/NVIDIA/framework-determinism".

\subsubsection{No Extra Tricks}
In our work, we strictly follow the training settings and implementation of DANet~\cite{cDualAttention} when comparing with other methods. 
Recently, many settings such as cosine decay learning rate or layer normalization has been widely used in computer vision to boost performance. 
In our experiments, we found they also worked well for our CAA, but we did not include them in this paper to conduct comparisons as fair as possible.

\subsection{The Detailed Architecture of CAA}

\begin{figure*}[!t]
	\centering
	\includegraphics[width=1.0\linewidth]{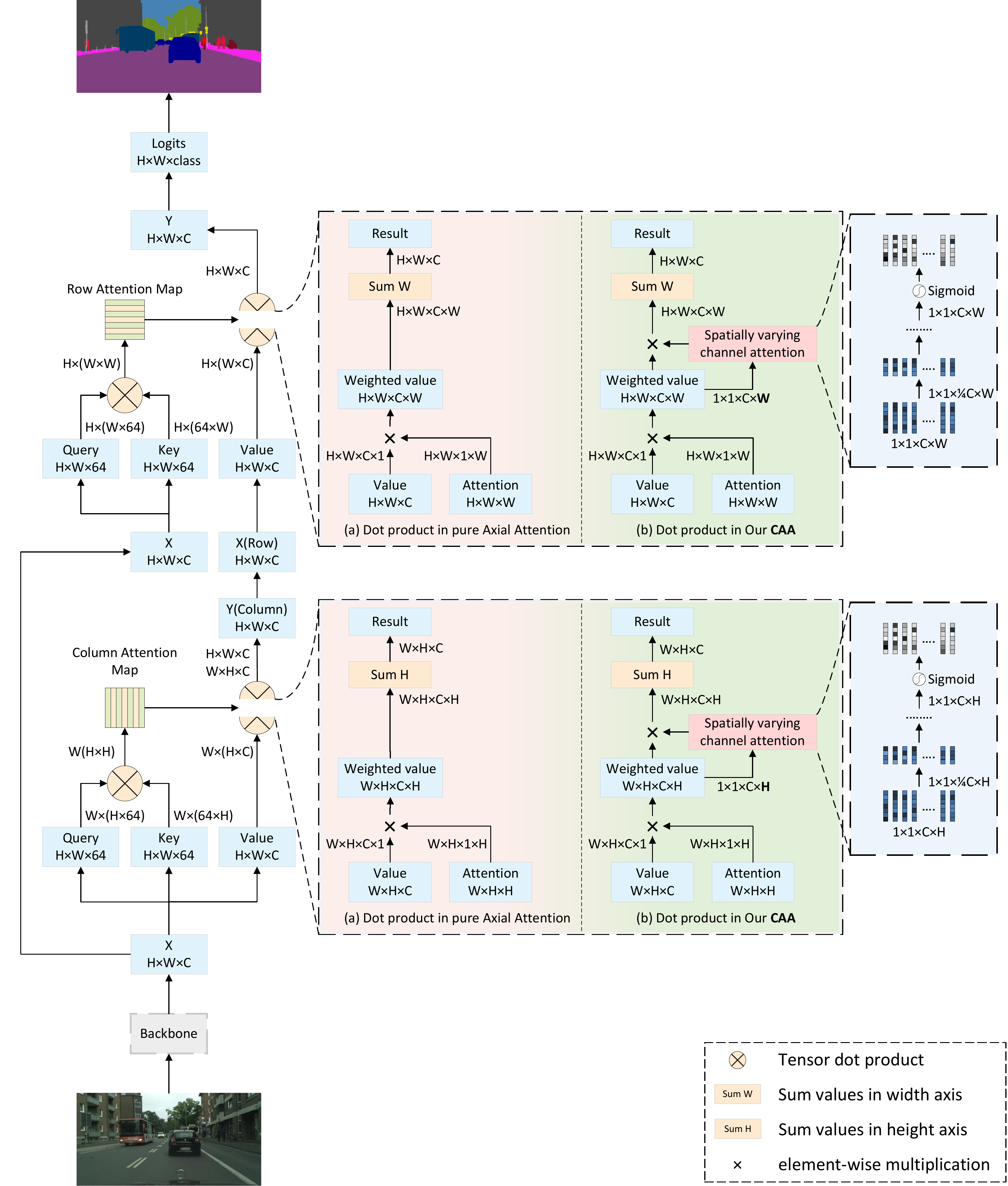}
	\caption{The detailed architecture of our CAA 
	}
	\label{figCAAFull}
\end{figure*}

Due to the page limits, we are only able to present a partial architecture (row attention) of our CAA in the main paper. 
In this section, we show the complete CAA architecture in Fig.~\ref{figCAAFull}. 
\clearpage
\clearpage

\bibliography{aaai22}

\end{document}